\documentclass[10pt,twocolumn,letterpaper]{article}
\let\SUP\textsuperscript
\usepackage{cvpr}             
\usepackage[accsupp]{axessibility} 

\usepackage{colortbl}
\usepackage[dvipsnames]{xcolor}

\usepackage[pagebackref,breaklinks,colorlinks]{hyperref}

\begin{document}
\title{ EfficientDreamer: High-Fidelity and Stable 3D Creation \\ via Orthogonal-view Diffusion Priors}

\author{Zhipeng Hu\SUP{1}$^\dagger$, Minda Zhao{\SUP{1}$^\dagger$}, Chaoyi Zhao\SUP{1}, Xinyue Liang\SUP{1}, Lincheng Li\SUP{1*}, \\ Zeng Zhao\SUP{1}, Changjie Fan\SUP{1}, Xiaowei Zhou\SUP{2}, Xin Yu\SUP{3} \\ \SUP{1}NetEase Fuxi AI Lab, \SUP{2}State Key Lab of CAD\&CG, Zhejiang University \SUP{3}University of Queensland}

\maketitle

\def\thefootnote{$\dagger$}\footnotetext{Equal contribution}
\def\thefootnote{*}\footnotetext{Corresponding author}

\begin{abstract}

While image diffusion models have made significant progress in text-driven 3D content creation, they often fail to accurately capture the intended meaning of text prompts, especially for view information. This limitation leads to the Janus problem, where multi-faced 3D models are generated under the guidance of such diffusion models.
In this paper, we propose a robust high-quality 3D content generation pipeline by exploiting orthogonal-view image guidance. First, we introduce a novel 2D diffusion model that generates an image consisting of four orthogonal-view sub-images based on the given text prompt. Then, the 3D content is created using this diffusion model. Notably, the generated orthogonal-view image provides strong geometric structure priors and thus improves 3D consistency. As a result, it effectively resolves the Janus problem and significantly enhances the quality of 3D content creation. Additionally, we present a 3D synthesis fusion network that can further improve the details of the generated 3D contents. Both quantitative and qualitative evaluations demonstrate that our method surpasses previous text-to-3D techniques. Project page: \url{https://efficientdreamer.github.io}.
\end{abstract}    
\vspace{-2.0em}
\section{Introduction}
\label{sec:intro}

\begin{figure}
\begin{center}
   \includegraphics[width=1.0\linewidth]{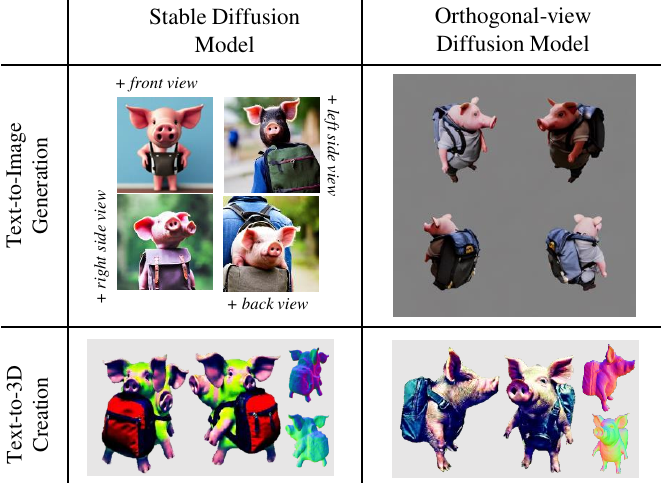}
\end{center}
\vspace{-1em}
   \caption{ The given prompt is \emph{A pig wearing a backpack, high quality.} {\bf Left:} The pre-trained Stable Diffusion model struggles to accurately generate images based on specific view instructions, thereby encountering the Janus problem in the text-driven 3D generation. {\bf Right:} \emph{EffcientDreamer} leverages our newly introduced orthogonal-view diffusion model, enabling the generation of 3D consistent images depicting the same scene from multiple orthogonal viewpoints.}
\label{fig:ablation}
\vspace{-1.0em}
\end{figure}


The field of generating photorealistic 2D images from simple text prompts is experiencing rapid growth, and recent advancements are largely attributed to the utilization of diffusion models. These advancements have led to the creation of models capable of producing images of exceptional quality \cite{ramesh2022hierarchical, rombach2022high, saharia2022photorealistic}. In contrast, generating high-quality 3D content from text prompts poses significantly greater challenges due to the much larger output space and the requirement for 3D consistency. Moreover, the scarcity of adequately large training pairs containing both text and 3D models further handicaps the development of effective models.


Early methods \cite{jain2022zero, lee2022understanding, mohammad2022clip} explore zero-shot text-guided 3D content creation by utilizing guidance from CLIP \cite{radford2021learning}. These approaches optimize the underlying 3D representations, such as NeRFs and meshes, to achieve high text-image alignment scores for all 2D renderings. However, these methods tend to generate 3D shapes with poor geometry and unsatisfactory appearance. To address these limitations, DreamFusion \cite{poole2022dreamfusion} showcases impressive capabilities in text-to-3D synthesis by leveraging a powerful pre-trained text-to-image diffusion model \cite{rombach2022high} as a strong image prior. A NeRF is trained to represent 3D models using novel Score Distillation Sampling (SDS) gradients \cite{poole2022dreamfusion} and view-dependent prompts. Subsequent works, such as TextMesh\cite{tsalicoglou2023textmesh}, Magic3D \cite{lin2023magic3d}, and ProlificDreamer \cite{wang2023prolificdreamer}, further improve the quality of the generated 3D contents.

While previous methods have demonstrated impressive results, a significant drawback is the Janus problem, which remains a major challenge. According to recent studies \cite{metzer2023latent, armandpour2023re}, the primary cause of the Janus problem lies in the inability of pre-trained 2D diffusion models to accurately interpret the view instructions specified in the text prompts. Even though explicit view instructions are given, the diffusion model may still generate images that do not align with these instructions. Since the stability of the 3D content creation process relies heavily on accurate guidance from the diffusion model, imprecise guidance can introduce instability and compromise the quality of the generated content.

To cope with these weaknesses, we propose {\emph{EfficientDreamer}}, which leverages a viewpoint-aware diffusion prior to create high-quality and consistent 3D contents. 
Our approach introduces a novel image diffusion model called {\bf orthogonal-view diffusion model}, which can generate a composite image comprising four sub-images that are mutually consistent. These sub-images, within a composite image, can be derived from arbitrary viewpoints while adhering to orthogonal views.
To achieve this, we maintain the architecture design of the 2D image diffusion model while adapting it to orthogonal-view image generation. We render the 3D models in a large-scale open-source 3D dataset called Objaverse \cite{deitke2023objaverse}, from orthogonal views, and assemble them to fine-tune the pre-trained 2D diffusion model. This process ensures the geometric consistency of our orthogonal-view diffusion model while preserving the capacity of 2D diffusion models to generate detailed and diverse high-quality objects that may not be present in the 3D datasets.

By leveraging our innovative orthogonal-view image diffusion model, we can effectively tackle the challenge of the multi-face Janus problem while generating consistent 3D models. Specifically, during each training step, we render the 3D representation (like Neus \cite{wang2021neus} or DMTet \cite{shen2021deep}) from different orthogonal viewpoints in a differentiable manner. These rendered images are then assembled into a composite image, which is further utilized to optimize the 3D representation via orthogonal-view score distillation.
While our orthogonal-view diffusion model, trained with the Objaverse 3D dataset, provides crucial geometric structure priors for 3D content creation, it is important to note that the scale of the 3D dataset is considerably smaller compared to the text-image training pairs. Additionally, the 3D assets in this dataset typically consist of common objects from the real world, having relatively simple topology and texture. 

Relying solely on such an orthogonal-view diffusion prior may lead to potential artifacts in the local appearance of the 3D contents. 
To address this issue, we also incorporate the original pre-trained 2D diffusion prior in the optimization of the 3D representation. In particular, we introduce a {\bf 3D synthesis fusion network} which integrates the orthogonal-view diffusion model with a pre-trained 2D diffusion model. We also present a dynamic 3D synthesis strategy to balance the guidance of the orthogonal-view diffusion prior and pre-trained 2D diffusion prior during the SDS optimization process. Initially, our focus is on generating 3D consistent geometric structures, where the high weight assigned to the orthogonal-view diffusion prior helps efficiently represent the geometry and alleviate the Janus problem. As the 3D objects become more complete, we gradually reduce the weight of orthogonal-view diffusion prior and increase the guidance from the original 2D diffusion prior. This approach allows for the generation of local texture details and refinement of local appearance.

Overall, our contributions are summarized as follows:
\begin{itemize}
\item We propose \emph{EfficientDreamer}, a novel approach for high-fidelity and robust 3D creation using orthogonal-view diffusion prior. This method effectively tackles the Janus problem and significantly enhances the stability of the generation process.

\item We introduce a novel image diffusion model that generates a composite image composed of four sub-images from orthogonal viewpoints. These orthogonal-view images serve as essential geometric structure priors, enabling the generation of 3D consistent geometric structures.


\item We present a novel text-driven 3D creation pipeline based on the orthogonal-view SDS loss. We introduce a 3D synthesis fusion network that combines the newly introduced orthogonal-view diffusion prior with the original 2D diffusion prior, aiming to ensure high-quality 3D creations.

\end{itemize}
\section{Related Work}
\subsection{3D Reconstruction with Neural Fields}

Traditional Multi-View Stereo (MVS) methods reconstruct 3D point clouds from predicted depth maps \cite{galliani2015massively, schonberger2016pixelwise, xu2019multi, zheng2014patchmatch}. Benefiting from deep learning, supervised MVS methods  \cite{gu2020cascade, wang2021patchmatchnet, yang2022non, yao2018mvsnet, yao2019recurrent} have shown impressive performance on benchmarks but require training on specific datasets  \cite{jensen2014large, yao2020blendedmvs}. These methods often struggle in less constrained scenarios and cannot easily integrate with other learning-based systems. 
On the other hand, in recent times, there have been significant advancements in neural fields that have demonstrated remarkable results across various tasks. As a famous neural volume rendering method, NeRF \cite{fridovich2022plenoxels, mildenhall2021nerf, muller2022instant, zhang2020nerf++} combines classical volumetric rendering with implicit function to render high-quality 2D images. However, directly extracting a mesh
from a NeRF representation is non-trivial \cite{tang2023delicate}.
Recent works have focused on improving the geometric network while establishing connections between density-based and surface-based representations. VolSDF \cite{yariv2021volume} models the volume density as Laplace Cumulative Distribution Function applied to an SDF representation.  Similar to VolSDF, NeuS \cite{wang2021neus} transforms the SDF field to the accumulated transmittance for volume rendering with the Logistic Cumulative Distribution Function and it is emphasized that the weight of the rendering should peak at the first intersection point from the outside to the inside. To this end, we adopt the NeuS representation for the convenience of mesh extraction.

\subsection{Text-to-image Generation}

In recent years, with the availability of extremely
large datasets of image-text pairs \cite{schuhmann2022laion},  significant progress has been made in text-to-image generation with diffusion models. A typical method is the Stable Diffusion model, which samples from a lower-resolution latent space and decodes latent into high-resolution images \cite{rombach2022high}. Such sampling operation in the latent greatly improves the efficiency of image generation. Another kind of method utilizes a cascade of super-resolution models \cite{balaji2022ediffi, saharia2022photorealistic}. These methods first generate a low-resolution image from a given text prompt and then enlarge the images with several super-resolution models. Recent works design special neural network structures to control diffusion models by adding extra conditions \cite{zhang2023adding}, it can learn task-specific input conditions and produce customized images. Due to the 
strong generative ability of pre-trained diffusion models, they are convenient to enable text-to-3D mesh synthesis.

\subsection{Text-to-3D Generation}

In recent years, text-to-3D has received significant attention due to the desire to create high-quality 3D content from simple semantics, such as text prompts. Early works attempt to use a CLIP objective to supervise the generation. CLIPMesh \cite{mohammad2022clip} deforms a 3D sphere using a CLIP loss to obtain a 3D mesh that fits the input prompt. Text2Mesh \cite{michel2022text2mesh} stylizes a 3D mesh by predicting color and local geometric details which conform to a target text prompt. Dreamfield \cite{jain2022zero} optimizes a NeRF from many camera views so that rendered images score highly with a target caption according to a pre-trained CLIP model. However, these methods tend to generate poor 3D shapes with unsatisfactory geometry as well as appearance.  

To overcome such limitations, DreamFusion \cite{poole2022dreamfusion} employs text-to-3D synthesis by utilizing a powerful pre-trained text-to-image diffusion model as a strong image prior. They optimize NeRF with their proposed Score Distillation Sampling (SDS) loss. Magic3D \cite{lin2023magic3d} extends such a method to a two-stage coarse-to-fine process, which utilizes DMTet for mesh representation. Fantasia3D \cite{chen2023fantasia3d} disentangles geometry and appearance into a two-stage optimization and introduces the spatially varying Bidirectional Reflectance Distribution Function (BRDF) into the text-to-3D task. TextMesh \cite{tsalicoglou2023textmesh} replaces NeRF with VolSDF for more accurate mesh expression. It proposes a novel multi-view consistent and mesh-conditioned re-texturing, enabling the generation of a photorealistic 3D mesh model. More recently, ProlificDreamer \cite{wang2023prolificdreamer} proposes Variational Score Distillation (VSD), a principled particle-based variational framework to promote the quality of created 3D models. 

\section{Methodology}

\begin{figure*}
\begin{center}
   \includegraphics[width=1\linewidth]{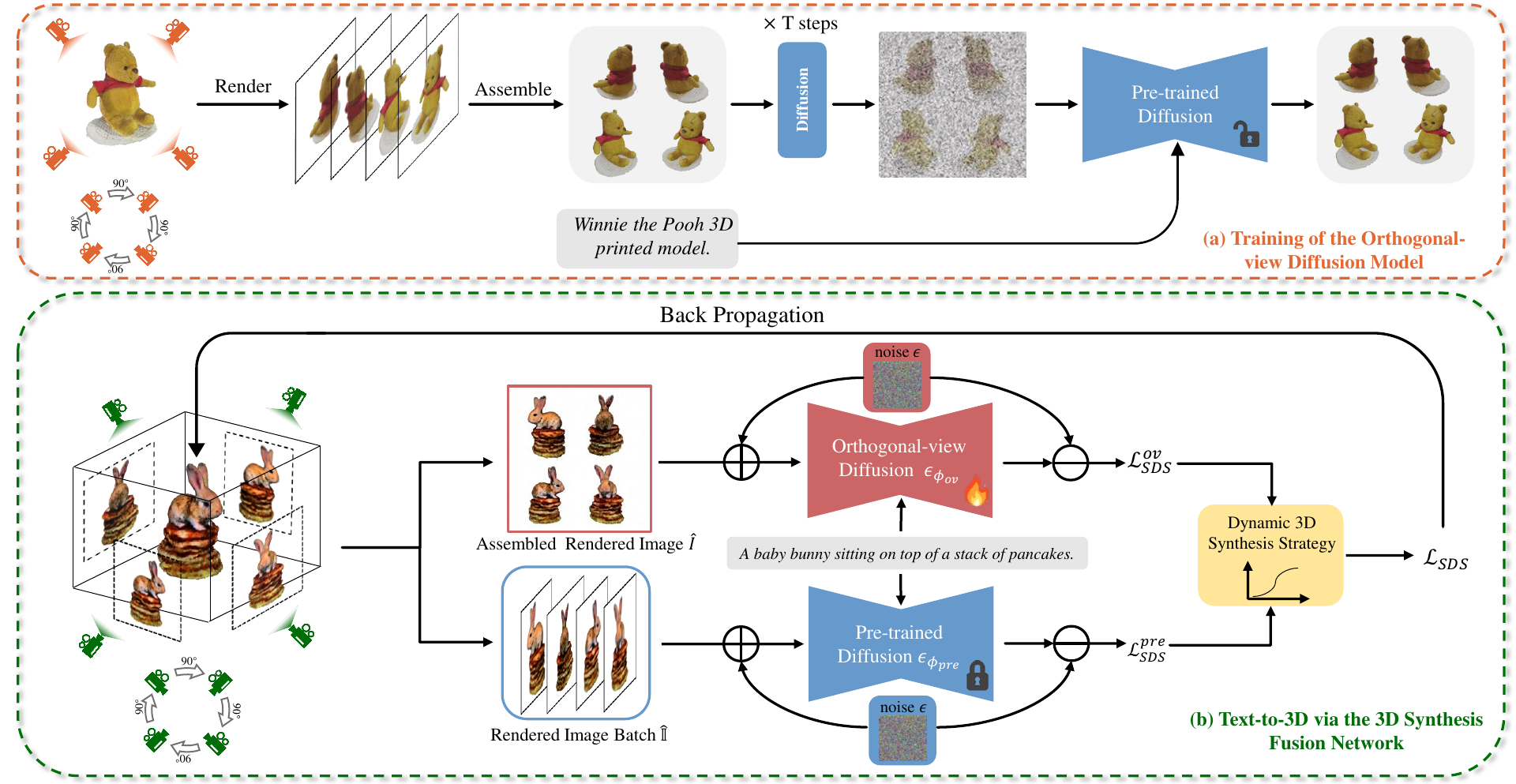}
\end{center}
\vspace{-1.0em}
   \caption{The overview of \emph{EfficientDreamer} involves two key steps. Firstly, we train an orthogonal-view diffusion model by rendering images from the Objaverse dataset. Secondly, we optimize the 3D scene representation by leveraging both the newly introduced orthogonal-view diffusion model and the pre-trained text-to-image diffusion model. To ensure high-fidelity and robust 3D creation, we employ a dynamic 3D synthesis strategy. }
   
\label{fig:pipeline}
\vspace{-1.0em}
\end{figure*}


Given a text prompt $y$, our goal is to create high-quality 3D mesh representations with photorealistic texture. To achieve this, we propose a novel orthogonal-view diffusion model, denoted as $\phi_{ov}$, which generates composite images consisting of four orthogonal-view sub-images. Subsequently, we design a 3D synthesis fusion network that combines the orthogonal-view diffusion model with a pre-trained diffusion model, denoted as $\phi_{pre}$, to enhance the quality of the generated 3D content.


\subsection{Orthogonal-view Diffusion Model}
\label{sec:Orthogonal-view}


Our objective is to generate a composite image in response to a given text prompt $y$ using a fine-tuned diffusion model. The composite image will consist of four sub-images from orthogonal viewpoints, arranged in a 2$\times$2 grid.

Common text-to-image diffusion models are trained on large-scale text-image pairs, enabling them to describe objects from various viewpoints. However, these models do not explicitly align corresponding parts of the object across viewpoints. Moreover, they tend to generate images of objects in canonical poses. Hence, it is essential to introduce a novel diffusion model that can generate representations of the same object from diverse viewpoints while maintaining 3D consistency. In this paper, we propose that showcasing the object from four orthogonal views is suitable as 
it provides essential geometric structure priors.

To obtain such a specific image diffusion model, we use the recently released Objaverse \cite{deitke2023objaverse} dataset for fine-tuning the commonly-used diffusion model, i.e., the Stable Diffusion model \cite{rombach2022high}. This dataset contains 800K+ 3D models created by 100K+ artists. The descriptions for each 3D model can be found in the Cap3D \cite{luo2023scalable} dataset, which is processed with BLIP2 \cite{li2023blip} and GPT4. We first filter some improper 3D models, like complex scene representation, textureless 3D models, and point clouds, and approximately 420K high-quality 3D models are left. We normalize all assets into a unit cube $[-0.5, 0.5]^3$. The radius (distance away from the center) is set at 1.8, with a field of view set at 35 degrees. For each 3D model, we uniformly render 12 images in the range of azimuth angle $\xi_{cam}\in (0^\circ, 360^\circ) $ and elevation angle list $\delta_{cam} = [0^\circ, 15^\circ, 30^\circ, 45^\circ]$. Thus, for each 3D asset, we render 48 RGBA images in 512$\times$512 resolution in total while its corresponding text prompt is denoted as $y$. We use the Stable Diffusion 2.1 as the original latent diffusion architecture with an encoder $\mathcal{E}$, a denoiser U-Net $\epsilon_{\theta}$, and a decoder $\mathcal{D}$. We utilize $\mathcal{E}$ and $\mathcal{D}$ with pre-trained weight and only fine-tune $\epsilon_{\theta}$ for generating composite images from orthogonal views. As illustrated in Fig. \ref{fig:pipeline}(a), for each training step, we first sample an elevation angle from the elevation angle list and then select four render images in this elevation angle, the rendered images are required to be 90 degrees apart from each other in azimuth angle. Then these four images are tiled on a 2×2 grid in a clockwise
rotation. To match the diffusion model, we resize this composite image into 512$\times$512 resolution,  which is denoted as $x$. At the diffusion time step $t\sim[1,1000]$, let $c(y)$ be the embedding of $y$, and $z_t$ be the noisy latent image by adding noise $\epsilon$ to a clean latent image $z$, we then solve for the following objective to fine-tune the model:
\begin{equation}
\mathop{min}\limits_{\theta} \mathbb{E}_{z\sim \mathcal{E} (x),t,\epsilon\sim \mathcal{N}(0,1)}||\epsilon-\epsilon_{\theta}(z_t,t,c(y))||^2.
\end{equation}

\begin{figure}
\begin{center}
   \includegraphics[width=1.0\linewidth]{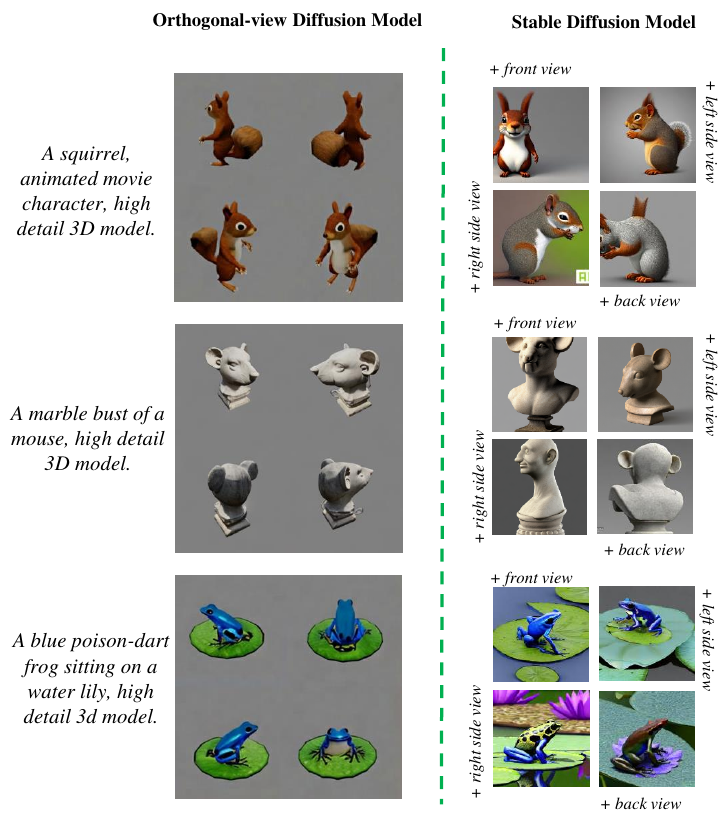}
\end{center}
   \vspace{-1.5em}
    \caption{Comparison between The orthogonal-view diffusion model and the pre-trained Stable Diffusion model with additional viewpoint instructions. The pre-trained Stable Diffusion model struggles to generate images based on specific view instructions, while the orthogonal-view diffusion model can generate composite images from orthogonal views.}
\label{fig:orthogonal}
\end{figure}

We find the trained diffusion model is capable of learning a generic mechanism to explicitly align the correspondence of an identical object from different viewpoints, even for objects never appearing in the training dataset. The comparison between the orthogonal-view diffusion model and the pre-trained diffusion model is depicted in Fig. \ref{fig:orthogonal}.

\subsection{Text-to-3D via the 3D Synthesis Fusion Network}
\label{sec:sds}

Once the orthogonal-view diffusion model is trained, we can generate our initial 3D model by training a neural distance field using the score distillation sampling strategy. The scene is represented as a differential render $g(\psi)$, where $\psi$ denotes the learnable parameters. Given a randomly sampled camera pose with azimuth angle $\xi_{cam}$,  elevation angle $\delta_{cam}$ and radius $r$, we fix $\delta_{cam}$ and $r$ and sample four camera poses by extending $\xi_{cam}$ to $[\xi_{cam}, \xi_{cam} + 90^\circ, \xi_{cam} + 180^\circ, \xi_{cam} + 270^\circ]$. Then we render images with these camera poses, which are denoted as $I_0$, $I_1$, $I_2$, $I_3$. We also tile them on a 2$\times$2 grid in a clockwise
rotation and resize the composite image into 512 $\times$ 512 resolution, which is denoted as $\hat{I}$. Then we sample random normal noise according to time step $t$ and add it to $\hat{I}$. 
The noisy images $\tilde{I}_t$, together with text embedding $c(y)$ are fed to our orthogonal-view diffusion model $\phi_{ov}$, which attempts to predict the noise $\epsilon$. 
The denoiser U-Net of $\phi_{ov}$ is denoted as $\epsilon_{\phi_{ov}}$ and the score function of noise estimation is denoted as $\epsilon_{\phi_{ov}}(\tilde{I}_t;t,c(y))$. This score function guides the direction of the gradient for updating the scene parameters, and the gradient is calculated by orthogonal-view Score Distillation Sampling (SDS):
\begin{equation}
\nabla_{\psi} \mathcal{L}_{SDS}^{ov}= \mathbb{E}_{t, \epsilon}[\omega(t)(\epsilon_{\phi_{ov}}(\tilde{I}_t;t,c(y)) - \epsilon) \frac{\partial \hat{I}}{\partial \psi}],
\end{equation}
where $\omega(t)$ denotes a weighting function. 

Motivated by \cite{huang2023dreamtime} and to enhance the stability of the training procedure, we decrease the time step (t) as training progresses, resulting in a gradual decline in the degree of added noise. As our training procedure supervises the 3D creation from four orthogonal views at each time step, the neural distance field can receive semantic structure guidance from different viewpoints. This fundamentally eliminates the possibility of creating multi-face problems.



Our orthogonal-view diffusion model, trained on the Objaverse 3D dataset, provides essential geometric structure priors for 3D creation. However, it is important to note that the scale of the 3D dataset is relatively small compared to the text-image training pairs. Additionally, the included 3D assets mainly consist of simple objects with uncomplicated topology and texture. Therefore, relying solely on the orthogonal-view diffusion prior may introduce potential artifacts in the local appearance of the generated 3D contents.

To resolve such a problem, we introduce a 3D synthesis fusion network which integrates the proposed orthogonal-view model with the original text-to-image diffusion model to guide the 3D model generation with a 3D synthesis dynamic strategy. As shown in Fig. \ref{fig:pipeline}(b), let $\epsilon_{\phi_{pre}}$ denote the denoiser U-Net of the pre-trained Stable Diffusion model, 
and rearrange the original rendered images {$I_0$, $I_1$, $I_2$, $I_3$} as an image batch $\hat{\mathbb{I}}$. $\tilde{\mathbb{I}}_t$ denotes the noisy result of $\hat{\mathbb{I}}$.  The corresponding SDS loss is formulated as:
\begin{equation}
\nabla_{\psi}\mathcal{L}_{SDS}^{pre}= \mathbb{E}_{t, \epsilon}[\omega(t)(\epsilon_{\phi_{pre}}(\tilde{\mathbb{I}}_t;t,c(y)) - \epsilon) \frac{\partial \hat{\mathbb{I}}}{\partial \psi}].
\end{equation}

We propose a new prior loss for the novel view supervision to combine both two priors:
\begin{equation}
\nabla_{\psi}\mathcal{L}_{SDS}= (1-(\frac{l}{L})^\lambda) \nabla_{\psi}\mathcal{L}_{SDS}^{ov} + (\frac{l}{L})^\lambda \nabla_{\psi}\mathcal{L}_{SDS}^{pre},
\end{equation}
where $l$ is the current iteration while $L$ is the total iterations. $\lambda$ is a hyperparameter to determine the strength of two diffusion priors, which is set as 1 in our experiments.


Initially, we prioritize generating 3D content with consistent geometric structures by emphasizing the orthogonal-view diffusion prior. As the generation progresses, we gradually shift the focus towards incorporating guidance from the original 2D diffusion prior to enhance local texture details and refine the overall appearance.

\subsection{Implementation Details}
\label{sec:scene_representation}

\noindent{\textbf{Orthogonal-view Diffusion Model Training:}}
Our model is implemented on the PyTorch platform with 8 NVIDIA Tesla A100 GPUs. We train the model utilizing AdamW optimizer with a learning rate of 1e-5. The whole training procedure takes 200K iterations with a batch size of 256, which takes about 5 days.

\noindent{\textbf{Text-to-3D Generation:}}
We employ a two-stage coarse-to-fine optimization process to generate high-quality 3D mesh representations with photorealistic textures from text prompts. In the coarse stage, we utilize NeuS \cite{wang2021neus} as the 3D scene representation, as density-based representations such as NeRF are not well-suited for extracting 3D geometry and obtaining meshes \cite{yariv2021volume}. The rendering resolution is set at 64x64, and the 3D representation is optimized for 5000 steps with 3D Synthesis Fusion Network. In the refinement stage, we employ DMTet \cite{shen2021deep} as the scene representation, enabling efficient rendering of textured meshes with differentiable rasterization at 512x512 resolutions. Additionally, we utilize SDS and VSD guidance \cite{wang2023prolificdreamer} to optimize the geometry and texture for another 5000 steps.
\section{Experiments}


In this section, we conduct extensive experiments to evaluate the effectiveness of our proposed method for text-to-3D content creation. We present qualitative and quantitative results, compare our method with state-of-the-art methods, perform a user study, and analyze the effectiveness of our proposed techniques through an ablation study.

\begin{figure*}
\begin{center}
   \includegraphics[width=1.0\linewidth]{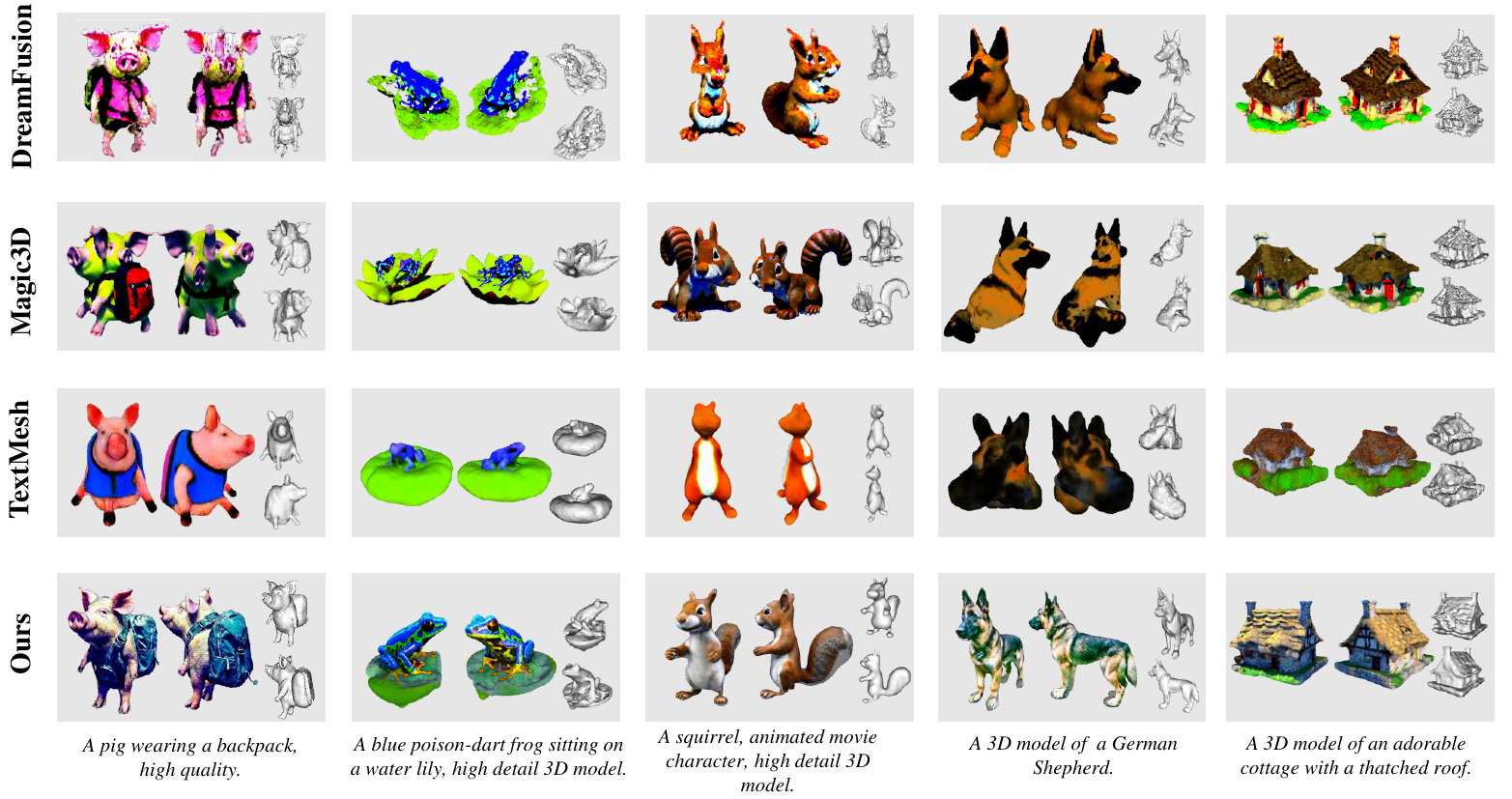}
\end{center}
\vspace{-1.0em}
   \caption{Comparison with other text-to-3D methods. We render each 3D model from two views. Our method outperforms other techniques by generating more high-fidelity 3D models without encountering the Janus problem.}
\label{fig:ablationl}
\vspace{-1.0em}
\end{figure*}

\subsection{Comparison with State-of-the-arts}
\label{sec:sota}
We present comprehensive experiments
to evaluate the effectiveness of our proposed method for text-to-3D content creation. The compared methods include three crucial Text-to-3D methods, i.e., DreamFusion \cite{poole2022dreamfusion}, Magic3D \cite{lin2023magic3d} and TextMesh \cite{tsalicoglou2023textmesh}. All these methods are implemented with a unified framework for 3D content creation \footnote{https://github.com/threestudio-project/threestudio}. As DreamFusion and TextMesh generate 3D contents with volume rendering of 64$\times$64 resolution, the generated 3D models may suffer from over-smoothness and lack of details. However, Magic3D can generate higher-quality 3D shapes with improved geometry and texture. Since our method also utilizes a coarse-to-fine two-stage optimization, we are able to create high-quality 3D mesh models. Furthermore, all of the compared methods may be confronted with the Janus problem as they rely on open-source pre-trained diffusion priors that do not enforce valid 3D consistency constraints. Our method can resolve the multi-face problem with our newly introduced orthogonal-view diffusion prior. Please
refer to the supplementary material for more comparison results.

We also evaluate our method using the CLIP score \cite{radford2021learning} and FID \cite{kynkaanniemi2022role} metrics. 
The CLIP score measures how well images rendered from the
generated objects correlate with the provided input text
prompt, while FID evaluates the quality and photorealistic appearance of the generated shapes. 
We render four images with fixed camera poses for each 3D model. The results are depicted in Table \ref{tab:metrics}, which demonstrates that our methods achieve better performance than the state-of-the-art methods.

\begin{table}[t]
\centering
\small{
\caption{Comparison with state-of-the-arts. Comparing
our method against the state-of-the-arts using CLIP score and FID. $\downarrow$ indicates the lower the better, and $\uparrow$ indicates the higher the better.}
\setlength{\tabcolsep}{2.0mm}
\begin{tabular}{lcccccc}
\toprule
Methods & CLIP $\uparrow$ & FID $\downarrow$  \\ \midrule
DreamFusion \cite{poole2022dreamfusion}  & 28.40 & 374.44  \\
Magic3D \cite{lin2023magic3d}  & 29.15 & 310.57  \\
TextMesh \cite{tsalicoglou2023textmesh} & 27.65 & 305.77  \\
\textbf{Ours}  & \textbf{30.33} & \textbf{284.98} \\
\bottomrule
\end{tabular}
\label{tab:metrics}
}
\vspace{-0.5em}
\end{table}

\subsection{Comparison with Perp-Neg}
\label{sec:perg-neg}

\begin{figure}
\begin{center}
   \includegraphics[width=1.0\linewidth]{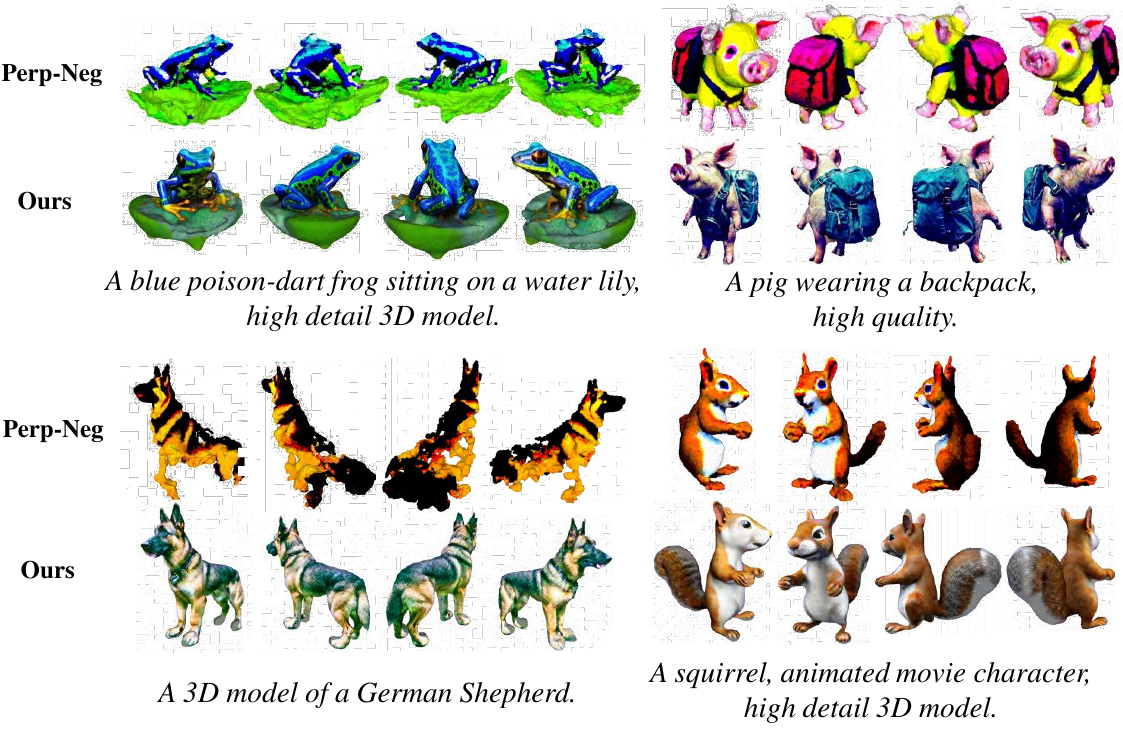}
\end{center}
\vspace{-1.5em}
   \caption{Comparison between the Perp-Neg method and our approach: Our method offers a more comprehensive solution to the Janus problem compared to Perp-Neg.}
\label{fig:perp-neg}
\end{figure}

In addition, we conduct a comparison between our method and another approach, Perp-Neg \cite{armandpour2023re}, which is designed to mitigate the Janus problem. We integrate Perp-Neg into DreamFusion and evaluate its performance against our method. The results, shown in Figure \ref{fig:perp-neg}, reveal that Perp-Neg can address the Janus problem to some extent. For instance, it is capable of generating 3D models with only one head and face. However, these results may exhibit unreasonable details such as three ears or five legs. In contrast, our method, which incorporates guidance from orthogonal views simultaneously, enables us to avoid such issues.

\begin{figure}
\begin{center}
   \includegraphics[width=1.0\linewidth]{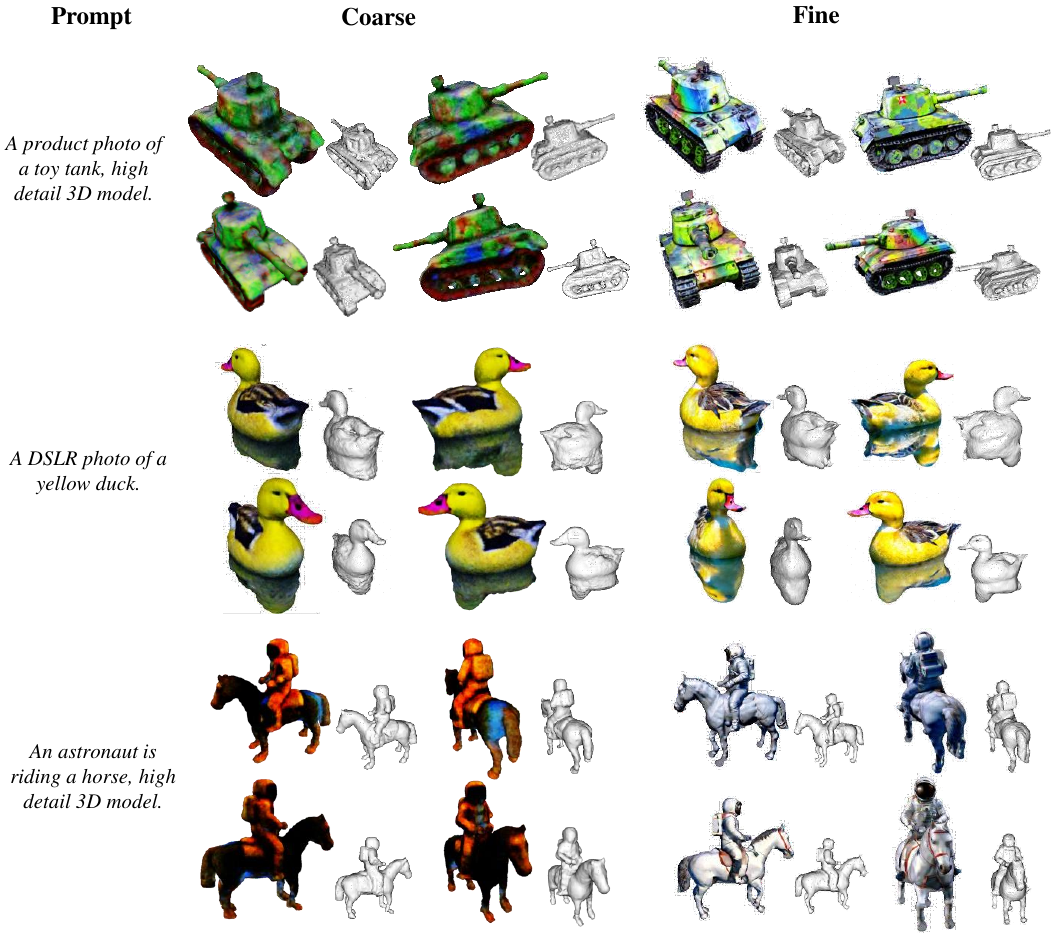}
\end{center}
\vspace{-1.5em}
   \caption{Coarse-to-fine two-stage optimization results. In the coarse stage, we get semantically reasonable 3D scene representation. In the fine stage,  we improve the visual quality of generated 3D assets.}
\label{fig:coarse-to-fine}
\vspace{-1.0em}
\end{figure}

\subsection{Coarse-to-fine Two-stage Optimization Results}
\label{sec:coarse-to-fine}

In Figure \ref{fig:coarse-to-fine}, we demonstrate the 3D contents generated through the coarse and fine stages of our optimization procedure. In the coarse stage, we achieve precise 3D representations for the given text prompts while effectively addressing the challenging Janus problem. This is accomplished with the guidance of the orthogonal-view diffusion model, which provides essential geometric structure priors for 3D creation. In the fine stage, we initialize the mesh representation with the results obtained from the coarse stage and proceed to optimize the geometry using SDS guidance. Additionally, we leverage VSD guidance to enhance the texture, ensuring that the generated 3D model is visually appealing and realistic. As a result, this process enables the generation of significantly higher-quality 3D models.

\subsection{User Study}
\label{sec:user-study}

We conduct a user study 
with Fuxi Youling Crowdsourcing \footnote{https://fuxi.163.com/solution/data}
for quantifying the subjective quality. Concretely, we evaluate the study on 22 prompts. At each evaluation time, the participants are shown the created 3D models by DreamFusion \cite{poole2022dreamfusion}, Magic3D \cite{lin2023magic3d}, TextMesh\cite{tsalicoglou2023textmesh} and our method,  and asked to rate them on a scale of 1 to 4. 
In total, we collect 1100 responses from 50 participants. The average rating scores for the different methods are shown in Table \ref{tab:user-study}. Notably, our method receives the highest average rating, indicating that it produces the most visually high-fidelity 3D models.
Furthermore, we calculate the preference ratios for each method based on the quality of the results. The analysis reveals that 84.27\% of the users consider the results of our method to be of higher quality than those generated by all competing methods. This finding further validates the effectiveness of our method in generating visually high-fidelity 3D models.


\begin{table}[t]
\centering
\small{
\caption{Results of our user study. The rating score represents the average rating results for different methods, while the preference indicates the ratio of user preference for the results of each method with higher quality.}

\setlength{\tabcolsep}{1.5mm}
\begin{tabular}{lcccccc}
\toprule
Methods & Rating Score $\uparrow$& Preference(\%) $\uparrow$  \\ \midrule
DreamFusion \cite{poole2022dreamfusion} & 2.09 & 4.55   \\
Magic3D \cite{lin2023magic3d}  & 2.44 & 6.27  \\
TextMesh \cite{tsalicoglou2023textmesh} & 1.73 & 4.91  \\
\textbf{Ours} & \textbf{3.74} & \textbf{84.27}  \\
\bottomrule
\end{tabular}
\label{tab:user-study}
}
\vspace{-1em}
\end{table}

\subsection{Ablation Study}
\label{sec:ablation}

\begin{figure}
\begin{center}
   \includegraphics[width=1.0\linewidth]{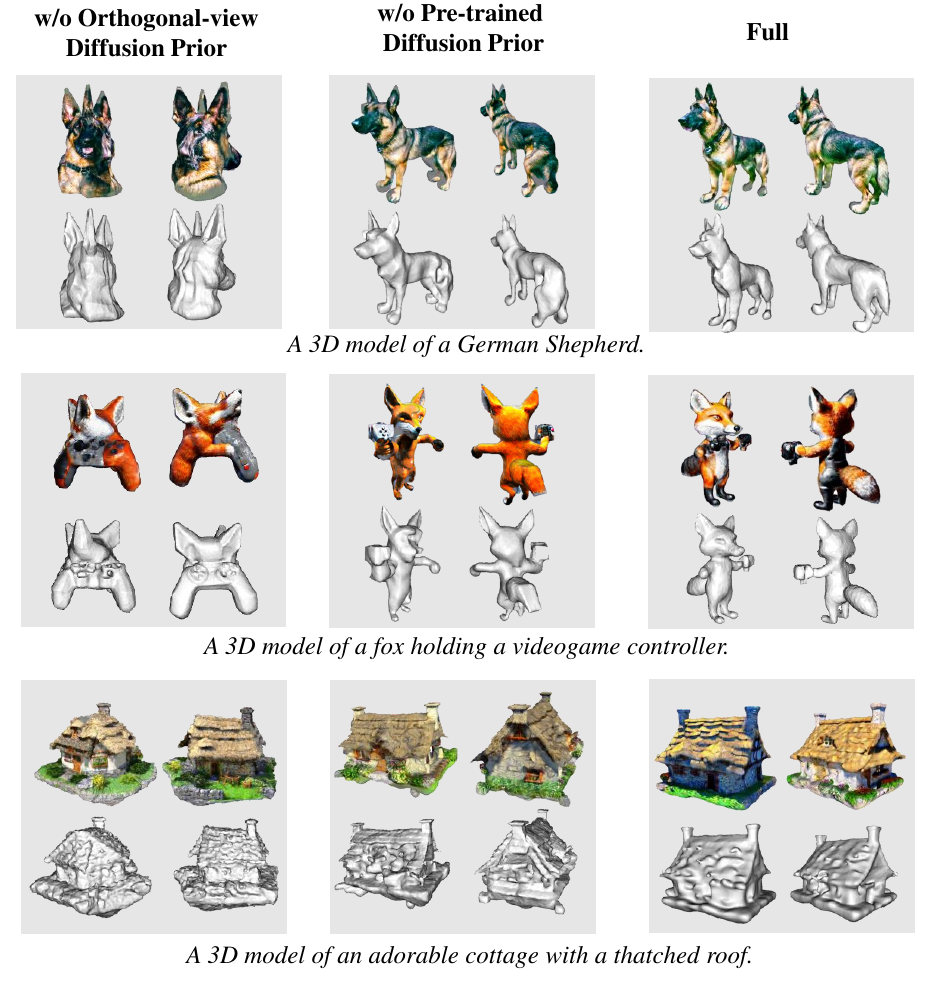}
\end{center}
\vspace{-2.0em}
   \caption{Ablation study on the impact of the  3D synthesis fusion network.}
\label{fig:ablation}
\vspace{-1.0em}
\end{figure}
\noindent{\textbf{Impact of the 3D synthesis Fusion Network:}}
We introduce a 3D synthesis fusion network with 
 the dynamic 3D synthesis strategy. Here we provide analysis and experiments to show its effectiveness.
Given several text prompts, we employ 3D creation with different diffusion model guidance.
The corresponding results are shown in Fig. \ref{fig:ablation}.
The first two columns show the results generated without the orthogonal-view diffusion model, which is equivalent to $\lambda=0$. Obvious Janus problems and inaccurate 3D representation can be found for the first two prompts. As for the creation of a cottage, we get similar shapes from different viewpoints. 
The next two columns demonstrate the result guided without the pre-trained diffusion model, which is equivalent to $\lambda=\infty$. While the orthogonal-view diffusion model can provide comprehensive structure constraints for 3D generation and restrain multi-face issues, we can see local distortion and unreasonable roughness. 
The last two columns demonstrate the results with our 3D synthesis fusion network, which means $\lambda=1$. We get complete and smooth meshes and eliminate the possible Janus problem. Quantitative results are depicted in Table \ref{tab:ablation}, which demonstrate the effectiveness of the 3D synthesis fusion network.

\begin{figure}
\begin{center}
   \includegraphics[width=1.0\linewidth]{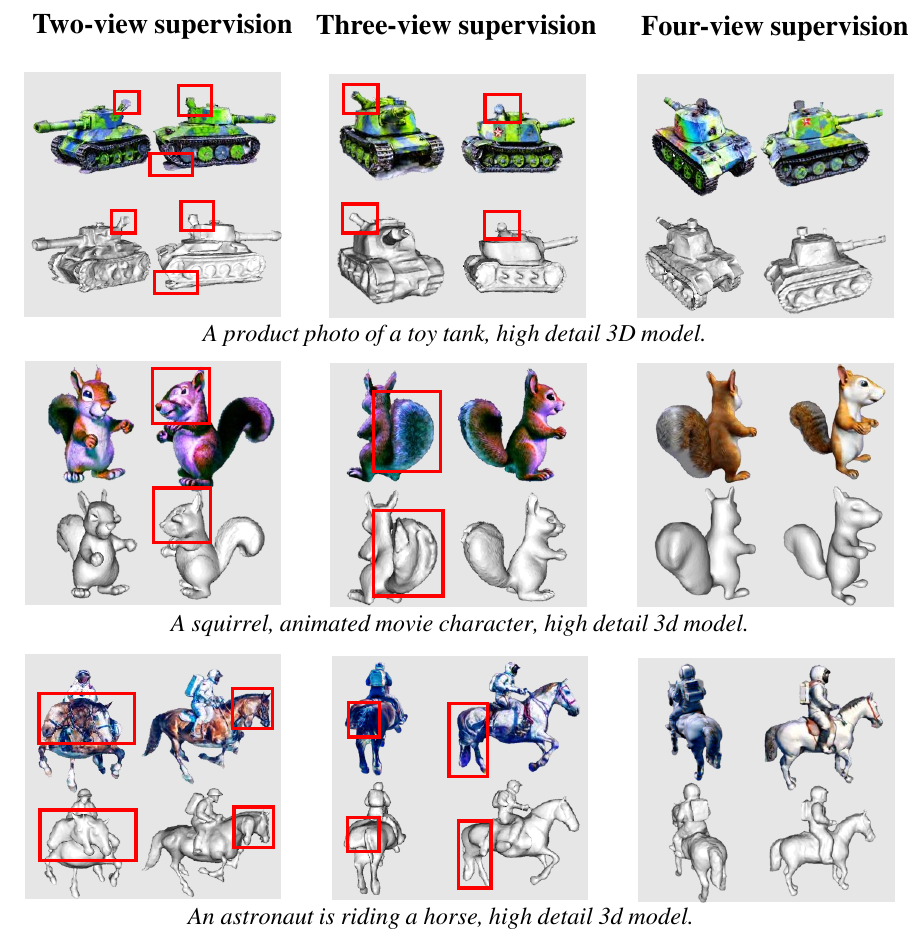}
\end{center}
\vspace{-2.0em}
   \caption{Ablation study on the impact of varying number of view supervision.}
\label{fig:two_three_ablation}
\end{figure}

\noindent{\textbf{Impact of Varying Number of View Supervision:}}
Here we present an ablation study on the effects of varying the number of view supervisions on our orthogonal-view diffusion model.  For two-view supervision, we randomly mask the opposite viewpoints in the rendered images during training, resulting in the optimization of the 3D representation using only two views. In the case of three-view supervision, we randomly mask one viewpoint, allowing the optimization of the 3D representation using three views. The corresponding results are shown in Figure \ref{fig:two_three_ablation}. Generally, supervising the 3D creation with our orthogonal-view diffusion model can mitigate Janus problems. However, creating 3D models with two-view supervision may still suffer from slight multi-face problems, such as two barrels in a tank or multiple faces in a squirrel and a horse. With three-view supervision, these problems are alleviated to some extent. Nevertheless, there are still local distortions in generated 3D models, such as unexpected sags or bulges. These issues can be addressed by using the intact four-view supervision of our orthogonal-view diffusion model.
Quantitative results are also presented in Table \ref{tab:ablation}, illustrating the impact of reducing the number of view supervision signals during training. It is evident that there is a decrease in performance observed in this scenario. This highlights the significance of optimizing a 3D representation using a set of four orthogonal-view images, as it provides sufficient geometric structure priors for ensuring consistent 3D content creation.

\begin{table}[t]
\centering
\small{
\caption{Ablation study results using CLIP and FID. $\downarrow$ indicates the lower the better, and $\uparrow$ indicates the higher the better.}
\begin{tabular}{c|ccc}
\hline
\multicolumn{4}{c}{\cellcolor[HTML]{EFEFEF}Impact of the 3D synthesis Fusion Network} \\ \hline
Metrics          & w/o Orthogonal-view   & w/o Pre-trained & Full        \\ \hline
CLIP $\uparrow$  & 31.13         & 32.05          & \textbf{32.55}   \\ 
FID  $\downarrow$  & 372.86        & 369.73         & \textbf{342.62}   \\ \hline
\hline
\multicolumn{4}{c}{\cellcolor[HTML]{EFEFEF} Impact of Varying Number of View Supervision} \\ \hline
Metrics          & Two-view            & Three-view     & Four-view   \\ \hline
CLIP $\uparrow$  & 30.92               & 30.48          & \textbf{31.47} \\ 
FID  $\downarrow$  & 399.58              & 373.04         & \textbf{345.72}\\ \hline

\end{tabular}
\vspace{-1em}
\label{tab:ablation}
}
\end{table}
\section{Conclusion}

In this paper, we present \emph{EfficientDreamer}, a novel approach for high-fidelity and robust 3D creation using orthogonal-view diffusion priors. We introduce a novel diffusion model that can create composite images comprising sub-images from different orthogonal viewpoints.  These composite images provide important geometric structure priors for 3D model generation by showcasing an object from multiple perspectives. We enhance the text-to-3D generation framework by leveraging the orthogonal-view diffusion model. Additionally, we introduce a 3D synthesis fusion network that integrates the orthogonal-view diffusion prior with the pre-trained diffusion prior. We also introduce a dynamic 3D synthesis strategy to balance the guidance of these two diffusion models during the SDS optimization. Our method effectively addresses the challenging Janus problem while ensuring the fidelity of the generated 3D models.
{
    \small
    \bibliographystyle{ieeenat_fullname}
    \bibliography{main}
}

\end{document}